# A multilayer perceptron-based fast sunlight assessment for the conceptual design of residential neighborhoods under Chinese policy

Can Jiang [a, b], Xiong Liang [a], Yu-cheng Zhou [a], Yong Tian [a], Shengli Xu [a], Jia-Rui Lin [b, *], Zhiliang Ma [b], Shiji Yang [c], Hao Zhou [d, e]

*a Glodon Company Limited, Beijing, China, 100193*

*b Department of Civil Engineering, Tsinghua University, Beijing, China, 100084*

*c Department of Building Science, Tsinghua University, Beijing, China, 100084*

*d Institute for Urban Governance and Sustainable Development, Tsinghua University, Beijing, China, 100084*

*e Key Laboratory of Eco Planning & Green Building, Ministry of Education (Tsinghua University), Beijing, 100084, China*

*\* Corresponding author: lin611@tsinghua.edu.cn*

**Abstract**

In Chinese building codes, it is required that residential buildings receive a minimum number of hours of natural, direct sunlight on a specified winter day, which represents the worst sunlight condition in a year. This requirement is a prerequisite for obtaining a building permit during the conceptual design of a residential project. Thus, officially sanctioned software is usually used to assess the sunlight performance of buildings. These software programs predict sunlight hours based on repeated shading calculations, which is time-consuming. This paper proposed a multilayer perceptron-based method, a one-stage prediction approach, which outputs a shading time interval caused by the inputted cuboid-form building. The sunlight hours of a site can be obtained by calculating the union of the sunlight time intervals (complement of shading time interval) of all the buildings. Three numerical experiments, i.e., horizontal level and slope analysis, and simulation-based optimization are carried out; the results show that the method reduces the computation time to 1/84~1/50 with 96.5%~98% accuracies. A residential neighborhood layout planning plug-in for Rhino 7/Grasshopper is also developed based on the proposed model. This paper indicates that deep learning techniques can be adopted to accelerate sunlight hour simulations at the conceptual design phase.

## 1. Introduction

The urban population of China has increased from 314 million to 883 million over the past three decades [1]. Rapid urban expansion has increased the demand for residential buildings [2], which caused investment in residential development to reach 11.1 trillion RMB, equivalent to 9.7% of China's GDP, in 2021 [3]. Local governments in China usually transfer neighborhood-scale land to real estate developers, who in turn build dozens of apartments that they sell to individual buyers [4]. To prevent profit-maximizing developers from excessively increasing the density of neighborhoods and to protect the health and well-being of residents [5, 6], "Standard for urban residential area planning and design" [7] and "Standard for assessment parameters of sunlight on building" [8] have been published to protect citizens' sunlight rights.

The sunlight policy in China is a national mandatory policy; construction plans that violate the requirements in standards [7, 8] cannot obtain building permits. Unlike many other countries that adopt annual or quarterly metrics [9, 10], Chinese policy-makers require that residential buildings receive sufficient sunlight on a specific winter day. The solar altitude on this day is the lowest, which represents one of the worst sunlight conditions for the entire year.

Enforcement of the Chinese sunlight policy is complicated and involves officially sanctioned software [11 - 13]. These simulators can not only predict the sunlight performance on sites but also allow simulation-based optimization of the layout of residential neighborhoods. The standard [8] also strictly defines the methodology of sunlight assessment, which relies on repeated shading calculation (SC). The recommended methodology is time-consuming (the computation time seems acceptable when conducting one sunlight simulation but is extremely long when conducting simulation-based optimization), and the cumulative-sky-based acceleration [14] is infeasible because the desired result is sunlight hours on a specific day.

To improve the efficiency of conducting simulation-based conceptual design, this paper adopts deep-learning technology for sunlight assessment under Chinese policy. The main contributions of this paper are proposing a one-stage method that skips repeated SC and only uses a multilayer perceptron (MLP) to predict sunlight hour heatmaps on sites with different layouts. Only using a MLP to predict neighborhood-scale sunlight heatmaps is extremely difficult, and the key to realize it is taking advantage of the characteristics of sunlight on a winter day, geometric invariance, and the idea of divide-and-conquer. Another contribution is that we propose an approach of training while data generation, and this approach contains mechanisms to improve the accuracy and generalization ability of the MLP model by avoiding the production of duplicate and inefficient training data. We also develop a plug-in on Rhino7/Grasshopper to optimize the conceptual layout planning of residential neighborhoods with the MLP model.

This paper mentioned ours and third-party software, and their roles are presented in Fig. 1. The remainder of the content is organized as follows. In Section 2, we review the techniques of SC-based and artificial neural network (ANN)-based building sunlight and daylight performance and solar irradiance simulation. Section 3 explain the features of Chinese sunlight assessment and conceptual residential neighborhood layout planning. In Section 4, we propose an SC-based sunlight analysis tool to generate data and train our MLP-based model; this tool is faster than others due to scenario-dependent code optimization, which is suitable for data generation with parametric modeling techniques. Section 5 compares our MLP model with our SC-based tool and Nvidia Optix [15] application programming interface (API); we have our own format for geometric data of residential neighborhoods and only have the automatic modeling interfaces to our tool, Nvidia Optix [15], and Open3D [16]. Section 6 concludes this study and discusses its applications and limitations. Appendix A lists the translation of the most important sunlight regulations in standards [7, 8]. Appendix B validated our SC-based tool with the Ladybug [17] plug-in, Open3D [16] API, Glodon Sunlight Analysis Software [13] (an officially sanctioned software developed by another team of our company) API, etc. Appendix C and D are proofs and technique details.

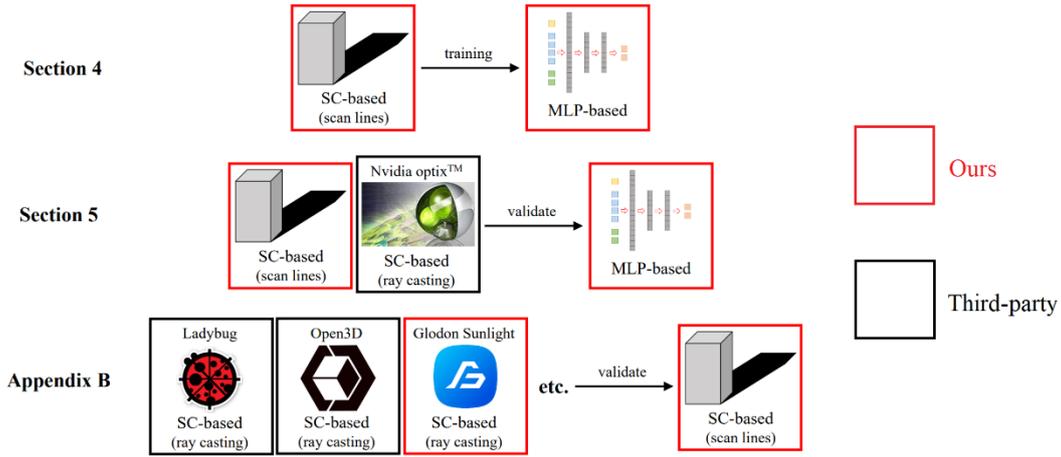

Fig. 1. Roles of ours and third-party software in the paper

## 2. Techniques review

### 2.1. SC-based simulation

SC techniques are critical to not only traditional sunlight hour prediction but also almost all physics-based lighting simulations. Sunlight, daylight and solar irradiance simulations are most popular in the field of building performance simulation; the latter two consider light rays not only directly from the sun but also diffused by clouds. The results of solar irradiance simulation are also important to building operational energy [18, 19] and gains of building-integrated photovoltaics [20, 21] simulations.

There are two kinds of SC algorithms: polygon clipping and pixel counting [22]. Polygon clipping algorithms [23, 24] output the sunlit fractions on surfaces based on the total sunlit areas, which is illustrated in Eq. (1), where the sunlit areas are calculated based on the Shoelace formula [25]. Although polygon clipping algorithms are used in the SC module of the famous energy simulator, i.e., EnergyPlus [26], these algorithms do not meet the demands of most building standards because they require heatmaps, i.e., sunlight hours of gridded sampling points.

$$\text{sunlit fraction} = 1 - \frac{\text{total shading area on the surface}}{\text{total area of the surface}} \quad (1)$$

Pixel counting algorithms can output heatmaps, where the value of a pixel represents whether the corresponding sampling point receives sunlight, daylight or irradiance. The pixel counting algorithm includes scan lines [27, 28], shadow maps [29], shadow volumes [30], and ray casting [31], and the difference among them is how to judge whether a sampling point is shaded. For the scan line algorithm, if a point is in the projected polygons of the shades, it is shaded; for the shadow map algorithm, if there is an obstruction in a similar position and closer to the light source than the point; for the shadow volume algorithm, if the point is in a shadow volume; and for the ray casting algorithm, if there is an obstruction in the ray between the point and the light source.

The scan line algorithm is the simplest and fastest algorithm, but it is only for SC on planes. The other three algorithms can conduct SC for 3D scenarios, and the ray casting algorithm is the most popular among them. The ray casting algorithm is a part of the ray tracing technique [32], which is widely used in real-time photorealistic rendering [33]. Since people realized that graphics processing units (GPUs) can significantly accelerate ray tracing [34], releasing APIs for hardware acceleration has become a new standard on commercial GPU cards, e.g., Optix [15] is the ray tracing API for Nvidia's GPUs.

Because these SC algorithms, especially scan line and ray casting, are simple and clear, there are still scholars who conduct building solar irradiance performance simulations based on self-developed SC modules [35]. We also developed SC-based tools to generate training data and validate the trained MLP-based model because data generation with mainstream business sunlight simulators is slow and inconvenient. Our SC-based sunlight simulator is faster than others because we adopt the scan line algorithm and conduct scenario-dependent code optimization, and its accuracy is also validated; the detailed records are in Appendix B.

## 2.2. ANN-based simulation

The classification of ANN [36] is shown in Fig. 2. The simplest type of ANN is the single-layer perceptron, which only has one hidden layer. If an ANN has more than one fully connected hidden layer, it is called a deep neural network (DNN) [37]. MLP [38] is the simplest type of DNN, which has more than one hidden layer. Both single and multilayer perceptrons are par2par, i.e., whose inputs and outputs are both values. The convolutional neural network (CNN) [39] is pix2par, i.e., whose inputs are images and outputs are values, which is widely used in image classification and objective detection (from images). Long Short-Term Memory (LSTM) [40] is seq2seq, i.e., whose inputs and outputs are sequences, which is widely used in Natural Language Processing. The generative adversarial network (GAN) [41] is pix2pix, i.e., whose inputs and outputs are both images, which is widely used in AI-generated content.

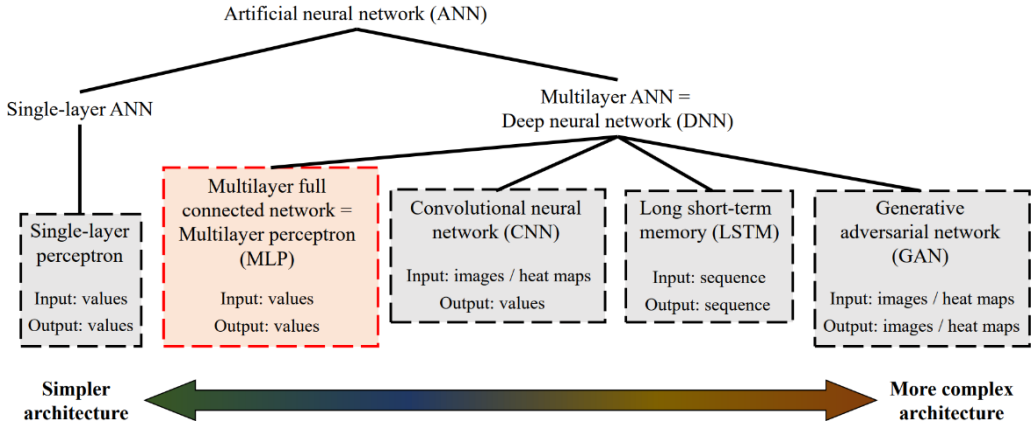

Fig. 2. classification of ANN

Research on ANN-based sunlight hour prediction is rare, but many scholars have predicted daylight and solar irradiance metrics on buildings with ANN models. Table 1 shows that these ANN-based building performance simulators have at least one of the following two shortcomings compared with our MLP-based model: (1) they are only applicable for calculation points with few fixed locations or a fixed scenario, and (2) they adopt more complex ANN architectures. Specifically, Kazanasmaz et al. [42] used a three-layer MLP to predict daylight illuminance on fixed points in a room with any dimensions and orientation; Lorenz et al. [43] only use a single-layer perceptron to predict daylight autonomy metric, but it can only predict for fixed points and different rooms need different ANNs; Wang et al. [44] proposed two-layer and three-layer MLPs to predict five building performance metrics (including one daylighting and two sunlight metrics), which are applicable for dynamic layouts of the site but fixed points; Chen et al. [45] used an MPL-based model to predict the equivalent obstruction angle (an important daylight metric) on a point on building façade, the model is applicable for different skylines but the relative location between the calculation point and the inputted skyline is fixed; Kristiansen et al. [46] predicted the heatmaps of annual illuminance in a specified room with a five-layer MLP; He et al. [47] used a GAN, a complex architecture, to predict heatmaps of the annual daylight metrics on residential floorplans; Han et al. [48] used a 3D CNN to predict solar radiation on building facades; Liu et al. [49] generate images for luminance-based analysis with a CNN-based model; Le-Thanh et al. [50] use a three-layer MLP model to produce heatmaps of useful daylight illuminance in rooms, but the number of neurons of their model is far more than ours.

Table 1 Comparison among existing ANN-based prediction models and our model

| Reference | Year | Predicted metric | Type of model | Dynamic points | Dynamic scenario |
|-----------|------|------------------|---------------|----------------|------------------|
| [42] | 2009 | daylight | MLP | × | √ |
| [43] | 2018 | daylight | single-layer perceptron | × | × |
| [44] | 2021 | sunlight and daylight | MLP | × | √ |
| [45] | 2021 | daylight | MLP | × | √ |
| [46] | 2022 | daylight | MLP | √ | × |
| [47] | 2021 | daylight | GAN | √ | √ |
| [48] | 2022 | solar irradiance | 3D CNN | √ | √ |
| [49] | 2020 | daylight | CNN | √ | √ |
| [50] | 2022 | daylight | MLP | √ | √ |
| ours | 2023 | sunlight | MLP | √ | √ |

## 3. Problem definition and analysis

### 3.1. Problem definition

The MLP-based model is used for sunlight assessment under the Chinese policy, and it can evaluate whether the sunlight hours of all residential buildings on a site satisfy the mandatory regulations of the standards [7, 8] and support optimizing the layout accordingly. The translation of the corresponding sunlight regulations in standards [7, 8] is listed in Appendix A.

Generally, the codes require that the "sunlight duration time" on the "reference positions" of residential buildings during the "period of effective sunlight" must be greater than or equal to the "minimum sunlight duration time". The "reference positions" correspond to the windowsill on the south façade of the residential building's first floor, and the standards [7, 8] define that the "reference positions" are 0.9 meters above the first floor's ground. Thus, architects usually calculate the heatmap of sunlight hours on the plane 0.9 meters above the ground of the site and check whether the values on the nearest points south of the south facades are greater than the "minimum sunlight duration time", as shown in Fig. 3 (a). The "period of effective sunlight" is a time interval on a specified winter day, which varies depending on the location of the site; it is 8:00 and 16:00 apparent solar time (AST) on January 20th, 2001, in Shanghai, and the sun trajectory of this period is shown in Fig. 3 (b).

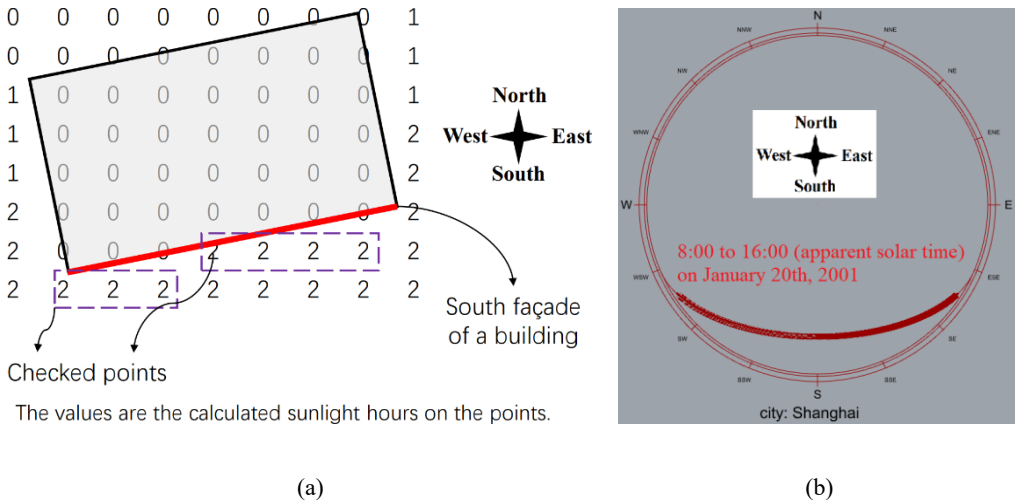

(a)                                        (b)

Fig. 3. Concepts of the Chinese sunlight assessment: (a) checking the "sunlight duration time" on the "reference positions"; (b) the sun trajectory of "period of effective sunlight"

The standard [8] also defines the process to calculate "sunlight duration time", and we generalize its pseudocode, which is illustrated in Table 2. The start and end times are needed, which are 8:00 and 16:00 (January 20th, 2001) in Shanghai. The recommended process needs to repeatedly conduct SC to calculate the ***sunlit*** matrix in Line 4, which is a Boolean matrix whose elements represent whether the sampling points are in the shadow (the values are equal to 0 if in the shaded area and 1 if in the sunlit area). It is obvious that both the accuracy and computation cost depend on the time step; both are lower if the time step is longer and higher if it is shorter. The standard [8] strongly recommends that the time step is equal to or shorter than 1 minute (a time step longer than 5 minutes is prohibited). The standard [8] explained that this is to ensure accuracy because a building may be overshadowed by multiple other buildings in the residential neighborhood, which causes a continuous sunshine interval of only a few minutes or less; a longer time step may lead to inaccurate simulation results. A 1-minute time step means 480 SC, which causes one sunlight assessment to take a few seconds, and conducting simulation-based optimization (which contains thousands of assessments) is extremely time-consuming. Moreover, the cumulative-sky-based method [14] cannot be used to accelerate the process because the sun trajectory from 8:00 to 16:00 covers only a very small part of the sky dome, as shown in Fig. 3 (b).

Table 2 Pseudocode for SC-based predicting sunlight hours

| Function sunlight-hours prediction(start time, end time, delta time): | |
|---|---|
| 1 | *time* ← start time, ***sunlight-hours ← 0*** |
| 2 | while *time* < end time: |
| 3 | *direction of sunlight* ← solar position prediction(*time*) |
| 4 | ***sunlit*** ← solar shading calculation(*direction of sunlight*) |
| 5 | ***sunlight-hours*** ←***sunlight-hours*** + delta time × ***sunlit*** |
| 6 | *time* ← *time* + delta time |
| 7 | Return ***sunlight-hours*** |

### 3.2. Problem analysis

The following three features of this problem make the MLP-based fast prediction approach possible: (1) a point is only shaded once by a building (we can take advantage of the idea of divide-and-conquer due to the fact); (2) many cases are equivalent due to geometric invariance, which simplifies the calculation; and (3) the time interval of a point shaded by a building is strongly related to their relative positions.

### 3.2.1. Usage of divide-and-conquer

Sunlight assessments in China are usually for obtaining the building permits of residential neighborhoods in the conceptual design phase, and architects prefer to use "boxes" (a box represents a residential building) for quick space planning [51]. Because all the buildings are cuboid-form and the sunlight simulation is for a single day, we find that any point on the site is only shaded once by a building; the proof is in Appendix C.1. We further proposed a fast union method to calculate the total shading hours of a point according to the start and end times of the point shaded by any of the buildings on the site. The pseudocode of the fast union method is illustrated in Table 3, and the proof of its correctness is introduced in Appendix C.2. ***ST*** and ***ET*** represent start and end times shaded, whose sizes are M × N, where M and N are the numbers of buildings and sampling points, respectively. $ST_{m,\,n}$ and $ET_{m,\,n}$ mean the start and end times of the *n*th point shaded by the *m*th building before being sorted in Line 1, and they mean the start and end times of the *n*th point shaded for the *m*th times after being sorted in Line 1. In Lines 2 to 4, the total shading hours equals the last end time ($ET_M$) minus the first start time ($ST_1$) of a point is shaded, then minus the total time of the points not be shaded between $ET_M$ and $ST_1$.

Table 3 Pseudocode for the fast union method

| | |
|---|---|
| | Function fast union method (***ST***, ***ET***): |
| 1 | ***ST*** ← sort in first dim(***ST***), ***ET*** ← sort in first dim(***ET***) |
| 2 | ***total shading hours*** ← $ET_M - ST_1$ |
| 3 | for $m$ in (1, 2, 3, ......, M − 1): |
| 4 |    ***total shading hours*** ← ***total shading hours*** − min(0, $ST_{m+1} - ET_m$) |
| 5 | Return ***total shading hours*** |

### 3.2.2. Usage of geometric invariance

We find that the time interval of a point shaded by a building is unchanged under two types of geometric invariance transformation, i.e., translation and scaling, as shown in Fig. 4. Fig. 4 (a) illustrates the translation equivalence; the time interval of point $(x_0, y_0, z_0)$ shaded by building #A equals point $(x_0, y_0, 0)$ caused by building #B if building #A is $z_0$ higher than building #B and their sections are the same. Fig. 4 (b) shows the scaling equivalence; the time interval of point $(x_0, y_0, 0)$ shaded by building #A equals point $(2x_0, 2y_0, 0)$ caused by building #B if the length, width, and height of building #B are twice those of building #A.

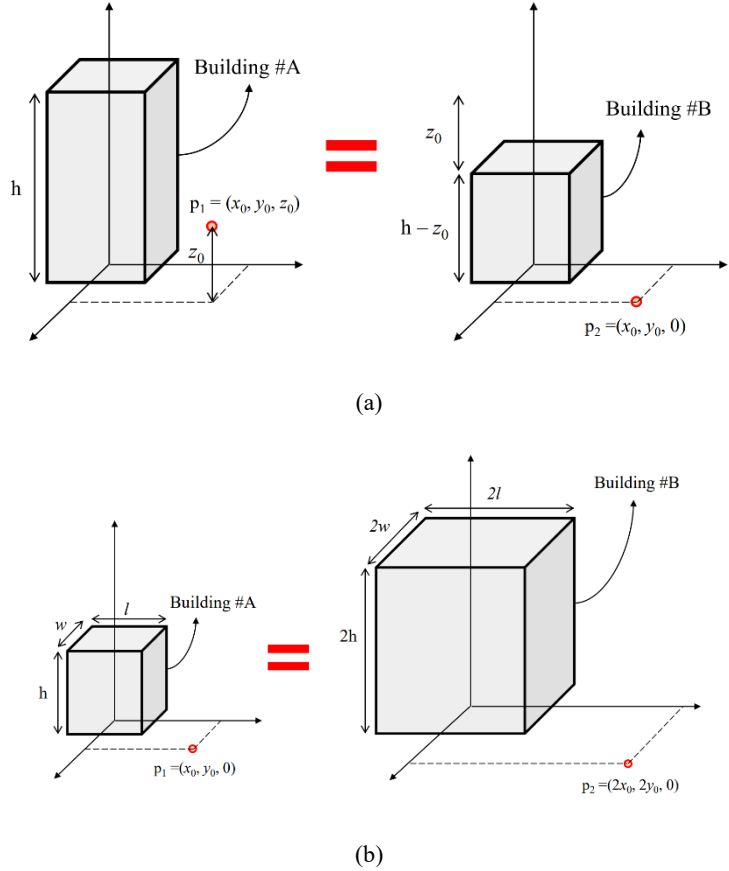

(a)

(b)

Fig. 4. Geometric invariance: (a) translation equivalence; (b) scaling equivalence

Because of the translation equivalence, calculating the sunlight hours on points with any height can be transformed into the problem of calculation for points on the ground. Because of the scaling equivalence, the absolute size of buildings and coordinates of points can be ignored, and we only need information on the relative size and location. These two types of geometric invariance simplify the training data generation and simulation via the MLP-based model.

### 3.2.3. Usage of the correlation of shading time and relative position

Appendix D.1 introduces an analytical method to calculate the shadow area of a building at any time (or sun location), and we can infer that the slenderer the building is, the less influence its section and orientation have on the shadow area. Thus, the time interval of a point shaded by a building mainly depends on its relative position if the building is not too flat. Fig. 5 (a) and (b) show the heatmaps of start and total time on the ground shaded by an example building (whose length, width, and height are 30, 21, and 48 meters, respectively) in Shanghai during the "period of effective sunlight". If we create a polar coordinate system whose origin is south of the building, it is obvious that the start and total time of a point, $(r, \theta)$, shaded by the building is strongly related to its polar coordinate. Because of the scaling equivalence (explained in Section 3.2.2), $r$ can be a dimensionless relative length.

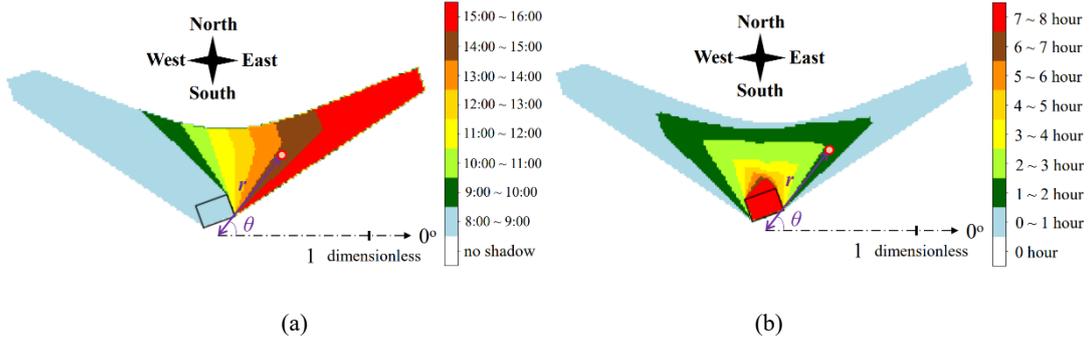

(a)                                         (b)

Fig. 5. Start and total time of points on the ground shaded by a building in Shanghai during the "period of effective sunlight": (a) heatmap of start time; (b) heatmap of total time

## 4. Methodology

Our approach is shown in Fig. 6, which includes data generation, training the MLP model and usage of the trained model. Fig. 6 (a) introduces the procedures of data generation while training. The architecture of the trained MLP model is illustrated in Section 4.1. The process of generating training data is illustrated in Section 4.2; we first generate the features via random parameterization and the mixed-scale and random-offset mechanisms; then, the features are fed to our SC-based tool to obtain the labels. Section 4.3 explains how to use a trained MLP to predict sunlight hours on sites, as shown in Fig. 6 (b).

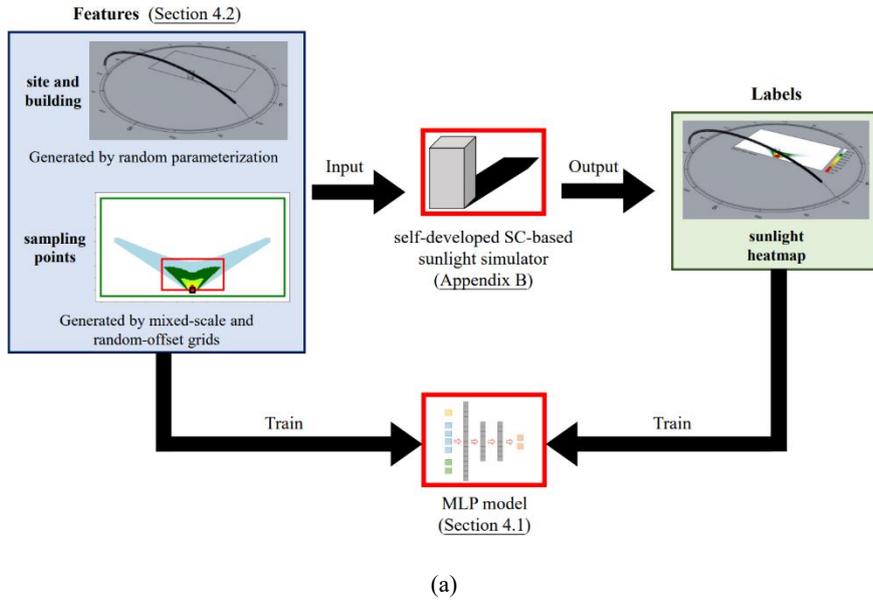

(a)

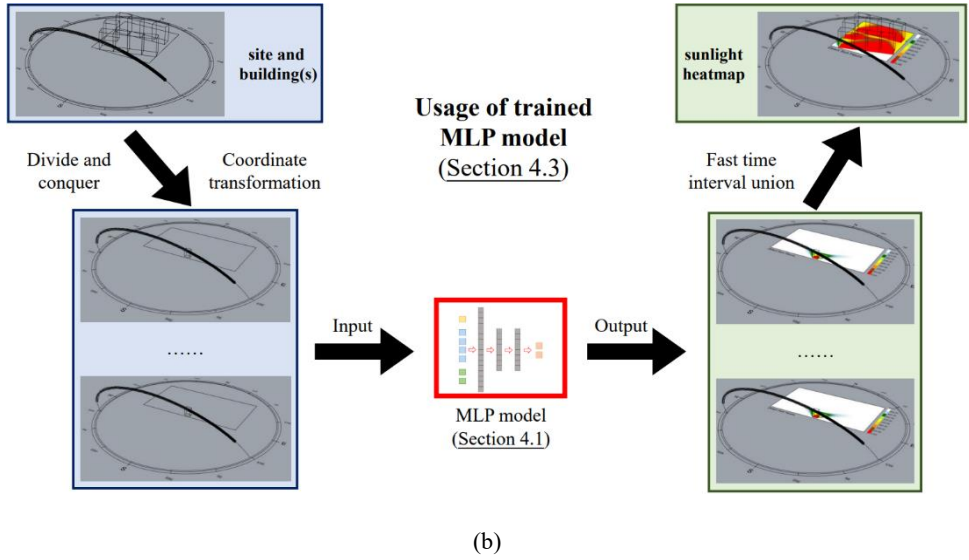

(b)

Fig. 6. Procedures of MLP-based sunlight assessment: (a) data generation and training; (b) utilization of the trained MLP model

### 4.1. MLP architecture design

#### 4.1.1. Inputs and outputs

An MLP model predicts the shading-time-interval caused by a cuboid-form building as Eq. (2) illustrated, and it directly calculates for sampling points on the ground, while points not on the ground can be handled via coordinate transformation because of the translation equivalence discussed in Section 3.2.2. Section 3.2.1 discussed that a point is only shaded once by a building during the "period of effective sunlight" defined by Chinese standards [7, 8]; thus, we represent the shading time interval by its start time and length, and its end time can be calculated by Eq. (3).

$$\text{shading time interval} = \text{MLP(site, building, sampling point)} \tag{2}$$

$$\textit{length of time interval} = \textit{end time} - \textit{start time} \tag{3}$$

Regarding the inputs, the latitude of the site, size and orientation of the building, and relative position between the sampling point and building are needed. The latitude of the site determines the solar trajectory during the "period of effective sunlight". Because of the effects of translation and scaling equivalence discussed in Section 3.2.2, we represent the size of the building by its relative size rather than the absolute size. The relative length, width, and height are calculated by Eqs. (4) to (7).

$$\textit{total size} = \textit{length} + \textit{width} + \textit{height} \tag{4}$$

$$\textit{relative length} = \textit{length} \div \textit{total size} \tag{5}$$

$$\textit{relative width} = \textit{width} \div \textit{total size} \tag{6}$$

$$\textit{relative height} = \textit{height} \div \textit{total size} \tag{7}$$

Section 3.2.3 illustrates that the shading time interval is strongly related to the relative position between the point and building in a polar coordinate system. However, sampling points of the sunlight heatmaps are from orthogonal grids in actual cases. We thus need both 2D local Cartesian and polar coordinate systems; points are sampled from the Cartesian coordinate system, while their polar coordinates are inputted to the MLP. We design these two coordinate systems as shown in Fig. 7. East is the $x$-direction and 0º-direction of the Cartesian and polar coordinate systems, respectively, and north is the $y$-direction and 90º-direction of the Cartesian and polar coordinate systems, respectively. The unit length in these two systems equals the total size in Eq. (4). The center of buildings must be in (0, 0.2) of the Cartesian system, i.e., (90º, 0.2) of the polar system.

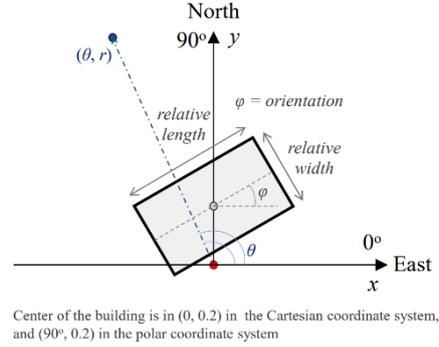

Fig. 7. Cartesian and polar coordinate systems that define the orientation of the building and the relative position between the building and sampling points

### 4.1.2. MLP architecture

Our MLP has only three hidden layers, as shown in Fig. 8, with sizes of 16, 8, and 8. All activation functions in these networks are rectified linear units (ReLUs), and none of the hidden layers contain batch normalization units.

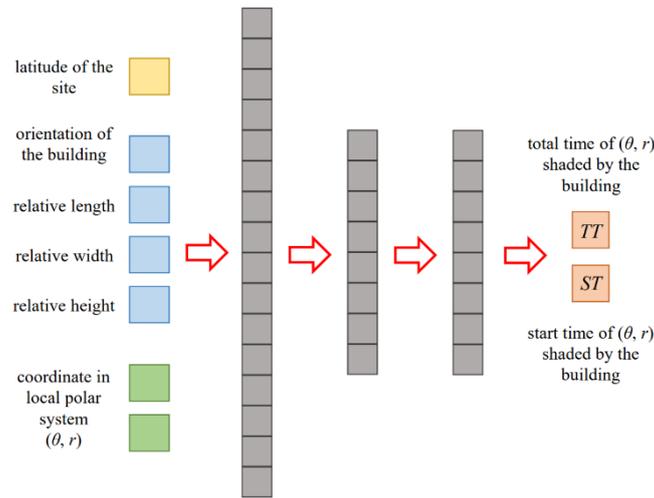

Fig. 8. Architecture of the MLP-based sunlight hour prediction model

### 4.2. Data generation and training

Training data include features and labels, which correspond to the inputs and outputs of the MLP, respectively. The features include the latitude of the site, size and orientation of the building, and polar coordinates of the sampling points. The former two are generated by randomly sampling from intervals, and the polar coordinates are generated with mixed-scale and random-offset grids. The labels are shading time intervals calculated by our SC-based simulator.

### 4.2.1. Random generation of latitude and size

Although our model can handle any possible scenarios in theory, the optimal mixed-scale grids depend on the latitude of the site and slenderness ratios of the building. We thus set the intervals of the latitude and slenderness ratios in Table 4; if the values out the intervals, the grids may need to be redesigned to ensure the efficiency and accuracy of training. The latitude interval is from 25 to 35 degrees north, which has approximately 850 million people, accounting for 60% of China's population [52]. We investigate the common slenderness ratios of residential buildings from our database and set the intervals accordingly.

Table 4 Interval of inputs for the MLP model

| Parameter | Latitude | Orientation | Length : Height | Width : Height | Length : Width |
|-----------|----------|-------------|-----------------|----------------|----------------|
| Interval | [25º N, 35º N] | [0º, 180º] | [1 : 4.33, 1 : 1] | [1 : 4.33, 1 : 1] | [3.33 : 1, 1 : 3.33] |

### 4.2.2. Generation with mixed-scale and random-offset grids

We design two mechanisms to sample points, i.e., mixed-scale and random-offset. Mixed-scale means sampling points form two grids with different scales. Fig. 9 shows an optimal grid size design under the conditions set in Table 4. The green box represents the coarse grid whose scale is 0.1, and its $x$ and $y$ coordinates (in the local Cartesian coordinate system) are between −3 and 3 and 0 and 3.2 (without random offset), respectively, because shadows of buildings must be in this area. The red box represents the fine grid whose scale is 0.05, and its $x$ and $y$ coordinates are between −1 and 1 and 0.2 and 1.2 (without random-offset), respectively, because the values of sunlight hours (or hours be shaded) for different sampling points have huge differences in this area.

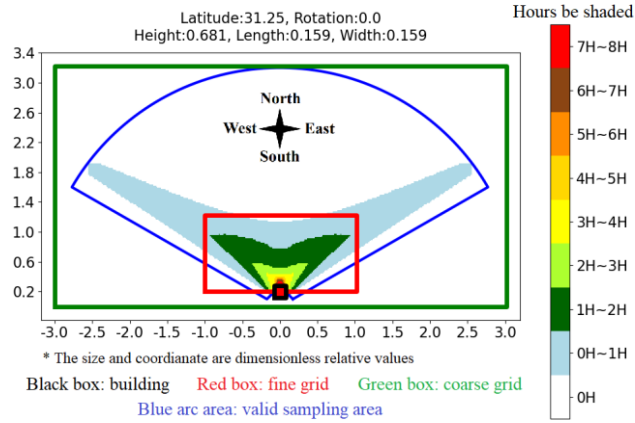

Fig. 9. Position of mixed-scale grid in the local Cartesian and coordinate system

Since we cannot calculate for sampling points that are not gridded, we design the random-offset mechanism to increase the diversity of positions of sampling points. The sampling points must be gridded for the following two reasons: (1) data generation is based on our SC-based sunlight simulator, and it uses a scan line algorithm [27, 28], which is only for gridded sampling points; and (2) the sampling points are usually gridded in real projects. We thus set that the start point of a grid is not fixed, i.e., the $x$ and $y$ coordinates of the start point of the coarse grid are randomly selected between −3.05 and −2.95 and −0.05 and 0.05, respectively, and between −1.025 and 0.975 and 0.175 and 0.225, respectively, for the fine grid.

Moreover, only points in the valid sampling area (blue arc area in Fig. 9) can be considered as training data, whose $\theta$ and $r$ coordinates (in the local polar coordinate system) are between 30º and 150° and 0.2 and 3.2, respectively. Shadows of buildings are almost always in the valid sampling area, and the proof is in Appendix C.3. We do not sample points in the region whose $r$ coordinates are between 0 and 0.2 because shadows of buildings are usually not in the region, while buildings are often in the region. Predicting sunlight hours for points under a building is not needed, and the values must be 0. Removing this region can also increase the continuity of the training data and further improve the accuracy of our model.

### 4.2.3. Training

We generate training data while training our model, and the amount of training data is not limited by storage space and can theoretically be infinite. The features of training data are generated in Section 4.2.1 and 4.2.2, and we then input the features to our SC-based sunlight simulator to obtain the labels. This simulator follows the methodology recommended by Standard [8] (described in Table 2), while the delta time is 1 minute.

We generate data for 256 scenarios at once. There are approximately 1774 sampling points for each scenario, approximately 1072 points from the coarse grid and approximately 702 points from the fine grid. These training data were randomly put into 64 batches.

The loss is defined by Eqs. (8) to (10),

$$\text{loss} = \text{total time loss} + \text{start time loss} \tag{8}$$

$$\text{total time loss} = (TT - TT_{\text{gt}})^2 \tag{9}$$

$$\text{start time loss} = \begin{cases} 2 \times (ST - ST_{\text{gt}})^2, \, TT_{\text{gt}} > 0 \\ 0, \, TT_{\text{gt}} = 0 \end{cases} \tag{10}$$

where $TT$ and $TT_{\text{gt}}$ are the total time of the sampling point shaded by the building predicted by the MLP and SC-based models, respectively, and $ST$ and $ST_{\text{gt}}$ are the start times of the sampling point shaded by the building, respectively. There are many sampling points not shaded by the building at all ($TT_{\text{gt}} = 0$), which do not have the $ST$s. The start-time losses are set to 0 for these points.

We trained the MLP-based model on a single thread of an i7-11700K @3.60 GHz CPU because training on a CPU is faster than training on a GPU for such a small neural network. We saved the weights and bias of the model every 10 updates. If the weighted average loss over the last 40 updates increases, we determine that the model converges and stop training. Fig. 10 shows the losses during the training process, which took 36216 seconds.

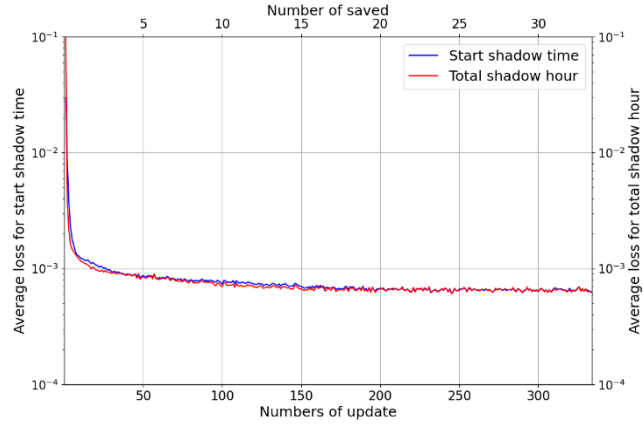

Fig. 10. Losses during the training process

### 4.3. Usage of trained MLP

#### 4.3.1. Calculation for points above or below ground

The MLP-based model directly predicts sunlight hours on ground, and it can also predict points above or below the ground. Because of the translation equivalence discussed in Section 3.2.2, the shading-time-interval on point $(\theta, r, z)$ caused by the building whose length, width, and height are $l$, $w$, and $h$ is same as shading-time-interval on point $(\theta, r, 0)$ caused by the building whose length, width, and height are $l$, $w$, and $h - z$. Coordinate transformation is necessary to preprocess the inputted parameters, and the details are explained in Appendix D.2.

#### 4.3.2. Calculation for sites with multiple buildings

As discussed in Section 3.2.1, we can handle the site with multiple buildings with the idea of divide and conquer. We need to call the MLP-based Model M times to simulate the site with M buildings. The pseudocode is presented in Table 5. M calls of the MLP model calculate the matrices of start and end times shaded, i.e., ***ST*** and ***ET***, whose sizes are M × N, where N is the number of sampling points on the site. In Line 5, the values of ***ST****m* and ***TT****m* are revised according to the rules described in Table 6, where ***ST****m* and ***TT****m* represent the start and total time shaded by the *m*th building, respectively. In Line 7, the fast union method proposed in Section 3.2.1 (its pseudocode is illustrated Table 3) calculates the heatmap of the sunlight hours based on ***ST*** and ***ET***. In Line 8, we find the points under the buildings and set their values to 0 (cannot receive sunlight at all), and the scan line algorithm is used to determine whether points are under buildings.

Table 5 Pseudocode for predicting sunlight hours on a site with the MLP-based model

| | Function sunlight-hours prediction(**MLP**, *building size*, *points on site*): |
|---|---|
| 1 | ***ST*** ← **0**, ***ET*** ← **0** |
| 2 | for *m* in (1, 2, 3, ……, M): |
| 3 | ***points on local system***, ***relative building size****m* ← coordinate transform(***building size****m*, ***points on site***) |
| 4 | ***ST****m*, ***TT****m* ← **MLP**(***relative building size****m*, ***points on local system***) |
| 5 | ***ST****m*, ***TT****m* ← post revise(***ST****m*, ***TT****m*) |
| 6 | ***ET****m* ← ***ST****m* + ***TT****m* |
| 7 | ***sunlight hours*** ← fast union method(***ST***, ***ET***) |
| 8 | ***sunlight hours***[***points under buildings***] = ***0*** |
| 9 | Return ***sunlight hours*** |

Table 6 Rules to revise *ST* and *TT*

| Rule #1 | if $TT < 0$: $TT$ ← 0 |
|---|---|
| Explain | Total time of a point be shaded cannot be less than 0 hour. |
| Rule #2 | if $TT > 8$: $TT$ ← 8 |
| Explain | Total time of a point be shaded cannot be more than whole time interval for analysis (8:00–16:00). |
| Rule #3 | if $ST < 0$: $ST$ ← 0 |
| Explain | Start time of a point be shaded cannot be earlier than 8:00 (AST). |
| Rule #4 | if $ST + TT > 8$: $ST$ ← $8 - TT$ |
| Explain | End time of a point be shaded cannot be later than 16:00 (AST). |

## 5. Results

As shown in Fig. 11, we validate the computation time and precision of the MLP-based model by comparing it with our SC-based sunlight simulator and Nvidia Optix [15]. Our SC-based tool is validated by many third-party software programs, e.g., Ladybug [17] and Open3D [16], and the details are in Appendix B, which proves that our SC-based tool is accurate and faster. Table 8 illustrates the setup of the test cases, all of which are in Shanghai, where the latitude is 31º 15' N. The analysis duration is 8:00~16:00 January 20th, 2001 (AST), because it is the "period of effective sunlight" in Shanghai according to Standard [7]. The position of the calculated points is 0.9 meters above the ground or slopes, which is the "reference position" defined by Standard [7].

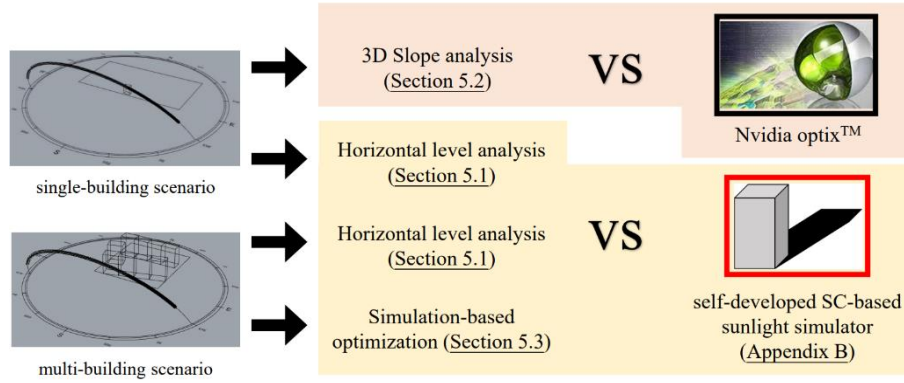

Fig. 11. Approach to validate our MLP model

Table 8 Setup of the test case

| Site | Duration | Position of calculated points |
|---|---|---|
| Shanghai (31° 15' N) | 8:00~16:00 January 20th, 2001 (AST) | 0.9 meters above the ground or slopes |

## 5.1. Horizontal level analysis

### 5.1.1. One building

We test 256 randomly generated scenarios and sampling points from the coarse grid with a 0.02-sampling interval (48000 sampling points for each scenario) and compare the results with our SC-based sunlight simulator. The average computation times and precisions are illustrated in Table 9. The MLP model significantly reduces the computation cost. Considering that it is possible to load the model only once and then use it an unlimited number of times, the MLP-based model can reduce the computation time by up to 1/50. The precision loss is acceptable, and the mean absolute errors (MAEs) for the total and start times are 1.375 and 5.717 minutes, respectively. Fig. 12 shows examples of the difference between the heatmaps of total and start time calculated by our SC-based and MLP-based models.

Table 9 Comparison between our MLP-based model and the SC-based tool

|  | Model loading time | Calculation time | Total time | Precision of *TT* | Precision of *ST* |
|---|---|---|---|---|---|
| (ours) SC tool |  | 0.308s | 0.308s | as Ground Truth | as Ground Truth |
| (ours) MLP | 0.001 s | 0.00606s | 0.00706s | MAE: 1.375 minutes | MAE: 5.717 minutes |

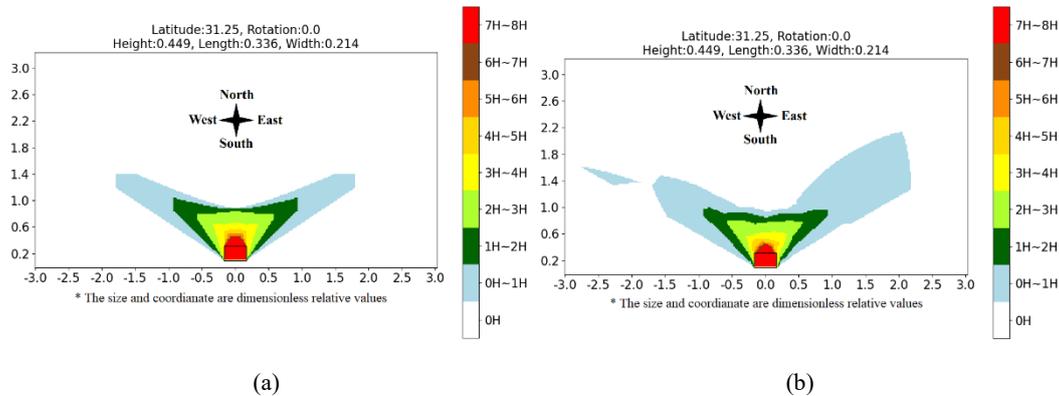

(a)                                    (b)

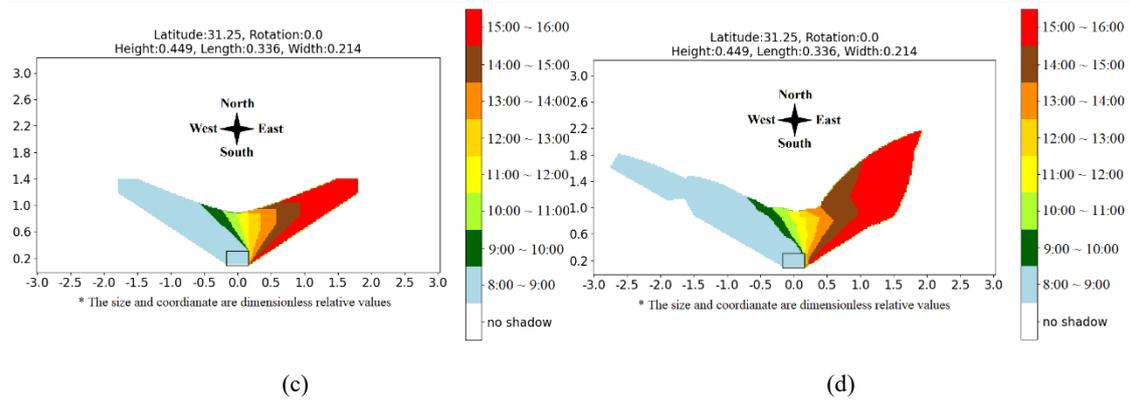

(c)          (d)

Fig. 12 Total and start time shaded by a cuboid-form building: total time calculated by our (a) SC-based tool and (b) MLP-based model; start time calculated by our (c) SC-based tool and (d) MLP-based model

### 5.1.2. Site with multiple buildings

We test the MLP-based model on a site with multiple buildings, which is a part of the early design of a residential neighborhood. The east–west and north–south lengths of the site are 150 and 160 meters, respectively, and the height of the buildings on the site is 45 meters. Fig. 13 shows the difference between the heatmaps of total time being shaded (the sunlight hours are 8 hours minus the total time being shaded) calculated by our SC-based and MLP-based model; these heatmaps are similar. Table 10 shows that the MLP-based model reduces the computation time by 1/77, while the MAE of the predicted total time to be shaded is 7.581 minutes, and the accuracy is 98.0%.

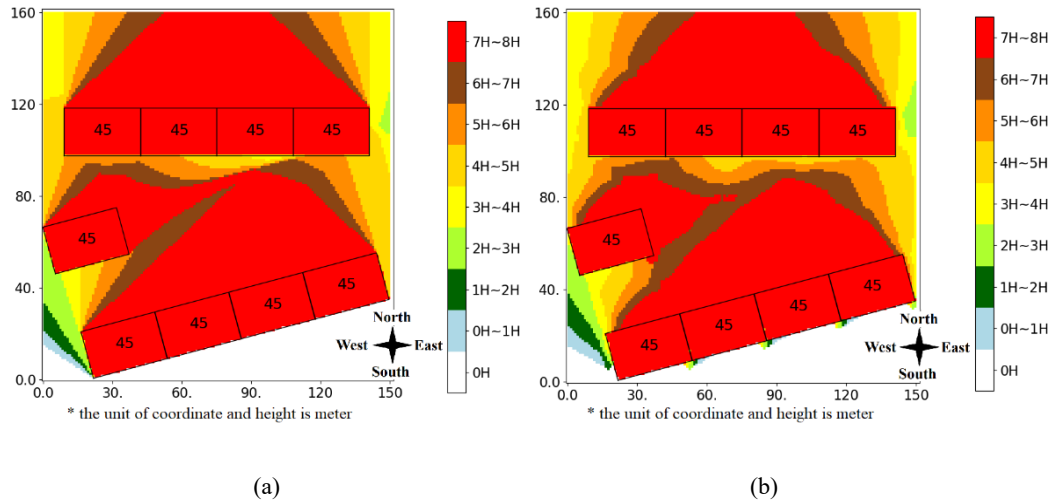

(a)          (b)

Fig. 13 Total time shaded on the site calculated by our (a) SC-based tool and (b) MLP-based model

Table 10 Comparison between our MLP-based model and the SC-based tool

|  | Model loading time | Calculation time | Total time | Precision of $TT$ | Accuracy |
|---|---|---|---|---|---|
| (ours) SC tool |  | 2.417s | 2.417s | as Ground Truth |  |
| (ours) MLP | 0.001 s | 0.0301 s | 0.0311 s | MAE: 7.581 minutes | 98.0% |

### 5.2. Slope analysis

This experiment tries to prove that the MLP-based model can predict sunlight hours for 3D points. Because we focus on sunlight assessment on the horizontal plane and our SC-based simulator is not available for 3D calculation, we did not collect the elevation data of real sites. We design a virtual site for this test, which is steeper than most real sites. The site is shown in Fig. 14, where a building is located in a plane whose elevation is 0 meters. The length, width, and height of the building are 20, 20, and 60 meters, respectively. The heights of all vertices of the slopes are shown in Fig. 14.

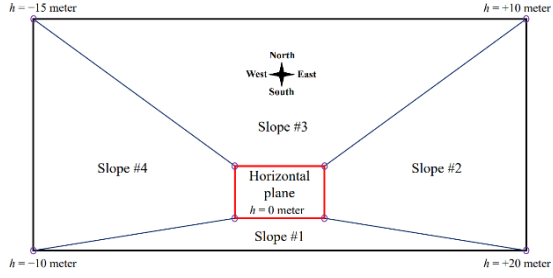

Fig. 14 Test site with slopes

We choose the Optix API of Nvidia to compare with our MLP-based model, which adopts a ray-casting-based approach for SC. This API is run on an Nvidia RTX 3080 GPU, while our model is run on a single thread of an i7-11700K @3.60 GHz CPU. The average computation times and precisions are illustrated in Table 11. The MLP model reduces the computation time by 1/84; the MAEs for the total and start times are 1.918 and 6.459 minutes, respectively. Fig. 15 shows the difference between the heatmaps of total and start time calculated by the Optix and the MLP-based model.

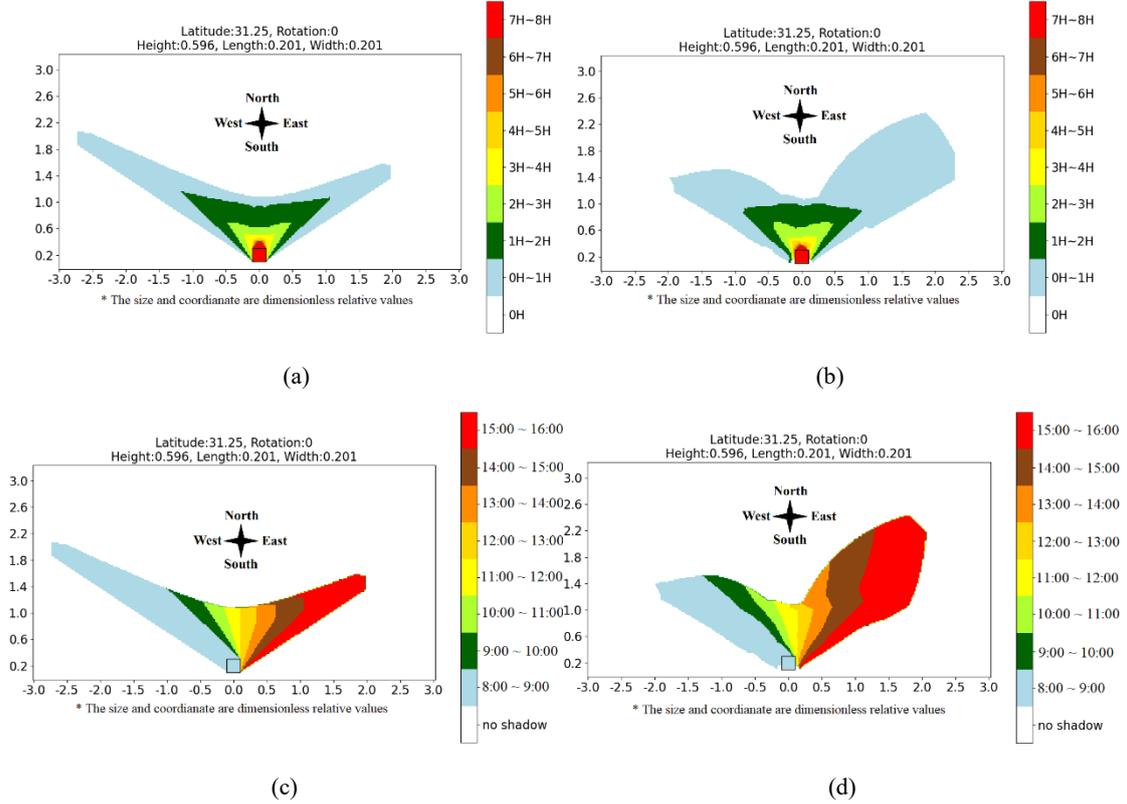

Fig. 15 Total and start time shaded by a cuboid-form building: total time calculated by (a) Optix of Nvidia and (b) our MLP-based model; start time calculated by (c) Optix of Nvidia and (d) our MLP-based model

Table 11 Comparison between our MLP-based model and Optix of Nvidia

|  | Model loading time | Calculation time | Total time | Precision of *TT* | Precision of *ST* |
|---|---|---|---|---|---|
| (Nvidia) Optix |  | 0.931 s | 0.931 s | as Ground Truth | as Ground Truth |
| (ours) MLP | 0.001 s | 0.0101 s | 0.0111 s | MAE: 1.918 minutes | MAE: 6.459 minutes |

## 5.3. Simulation-based optimization

The MLP-based model can significantly decrease the computation cost with little precision loss, as shown in Section 5.1 and 5.2. This section investigates whether the small precision loss affects the application of our model as a proxy model. Our MLP-based model is tested in a case of simulation-based optimization, and its outputs contribute to the objective function. The decision variables are the location, orientation, and size of a building, as shown in Fig. 16, and the purpose is to maximize the total floor area (or volume) of the building without violating the mandatory sunlight regulation [7].

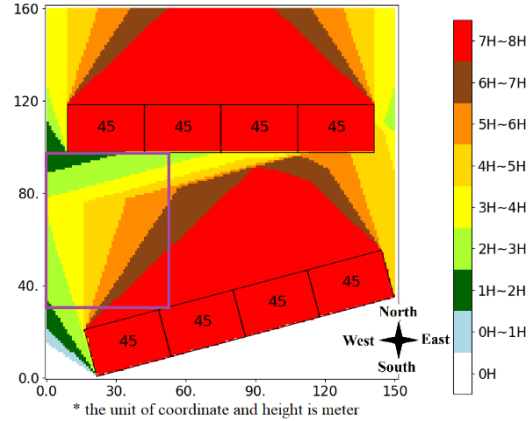

Fig. 16 Optimization problem: placing a building inside the purple box

Table 12 illustrates the ranges of the decision variables; the building is located by its westernmost and southernmost points, and its size and orientation are also determined. The objective function is defined in Eq. (11), whose details are illustrated in Appendix D.3. We use our MLP-based and SC-based models to calculate the sunlight hour score as a comparison. A genetic algorithm [53] is adopted to solve the problem, where the number of generations is 512, and the population of each generation is 64.

Table 12 Range of decision variables in the optimization problem

| Variable | $x$-coordinate (westernmost point) | $y$-coordinate (southernmost point) | Length | Width | Height | Orientation |
|---|---|---|---|---|---|---|
| Minimum | 0 m | 30 m | 21 m | 18 m | 30 m | -30º |
| Maximum | 7.5 m | 80 m | 42 m | 30 m | 60 m | 30º |
| Interval | 0.5 m | 0.5 m | 3 m | 3 m | 3 m | 5º |

$$\text{objective score} = \text{total floor area score} \times \text{direct-sunlight-hour score} \tag{11}$$

We first solve the problem 10 times with the MLP-based objective function, and the 10 solutions are exactly the same; we then solve it with the SC-based objective function, and the 10 solutions are exactly the same as well. Table 13 illustrates the solutions and corresponding scores and the average computation time. The error of the objective score calculated based on the MLP model is only 3.5% (the score calculated by the SC-based tool is considered as a ground truth), but the average computation time is reduced by 1/54. Fig. 17 shows that both of the solutions of the SC-based and MLP-based objective functions satisfy the sunlight regulations required by Standard [7] (the heatmaps are both calculated by the SC-based tool).

Table 13 Comparison between our MLP-based model and the SC-based tool

|  | Computation time | Solution | Score (MLP) | Score (Scan line) | Accuracy |
|---|---|---|---|---|---|
| (ours) SC tool | 7261.3 s | [0m, 42 m, 27 m, 30 m, 60 m, 0º] |  | 2.557 |  |
| (ours) MLP | 135.5 s | [0m, 42 m, 30 m, 30 m, 60 m, 0º] | 2.615 | 2.526 | 96.5% |

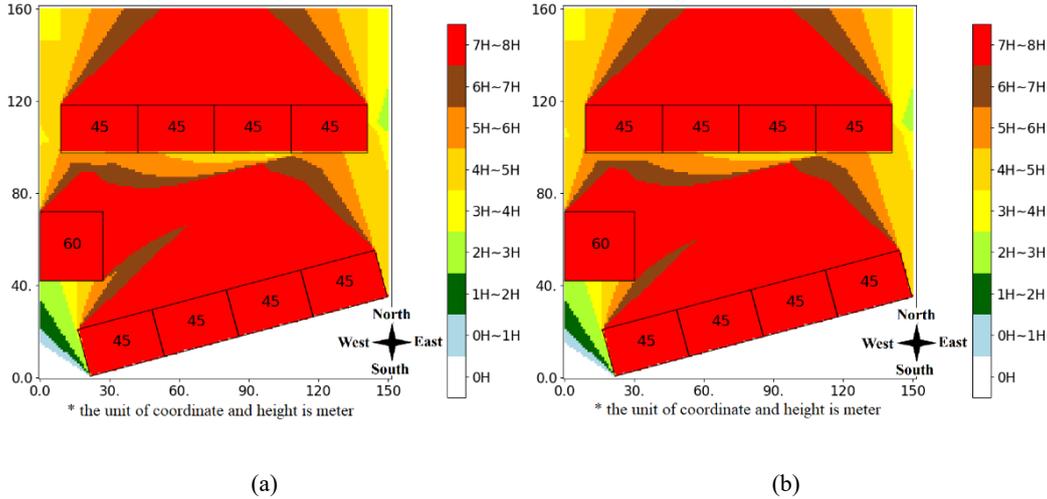

Fig. 17 Solutions of the optimization problem: (a) SC-based objective function, (b) MLP-based objective function

## 6. Conclusion and Discussion

### 6.1. Conclusion

Buildings receiving enough hours of sunlight is a mandatory regulation for Chinese residential neighborhoods design. Chinese architects and city planners use officially sanctioned solar simulators to conduct simulations and simulation-based optimizations to ensure that their designs do not violate the standards. These simulators rely on repeated solar shading calculations (SC), which is time-consuming, especially for simulation-based optimization.

To avoid repeated SC, this paper proposes a one-stage method to predict sunlight hours under Chinese policy. We take advantage of the following three features of Chinese sunlight assessment: (1) a point is only shaded once by a building; (2) translation and scaling equivalence; and (3) strong correlation between the shading time and relative position. A simple MLP network is designed to predict the heatmaps of the sunlight time interval (complement of shading time interval) caused by a single building. With coordinate transformation and our proposed fast time-interval union method, sites with multiple buildings can also be simulated.

The MLP model is trained by our proposed SC-based sunlight simulator, whose speed and accuracy are validated by many other third-party simulators. mixed-scale and random-offset grid mechanisms, and random parameterization are proposed to generate the features of training data, and the features are inputted to the SC-based simulator to obtain the labels.

The computation time and precision of the MLP-based model are validated on the horizontal level, slope analysis, and simulation-based optimization. Regarding the horizontal level analysis, sites with a single building are first tested; the computation time is reduced by 1/50, and the MAE for the total time to be shaded is 1.375 minutes. A test on a real design of a site with multiple buildings is then conducted; the computation time is reduced by 1/77, and the MAE for the total time to be shaded is 7.581 minutes (98% accuracy). Regarding the slope analysis, a test is carried out on a virtual site with four slopes; the computation time is reduced by 1/84, and the MAE for the total time to be shaded is 1.918 minutes. Regarding the simulation-based optimization, the MLP model is used to calculate the objective score, whose error is only 3.5% but reduces the computation time by 1/54, and the solution is almost exactly correct if the proxy objective function is used based on the MLP model.

Our model proves that deep learning and data-driven techniques can be adopted during sunlight hours and accelerate simulations, which will improve the efficiency for architects and city planners and ultimately contribute to improving the performance of buildings. The MLP model was used in practice; a Rhino 7/Grasshopper plug-in was developed based on it. As shown in Fig. 18, this plug-in is for automatic residential neighborhood layout planning, which places a row of buildings while checking the compliance of the sunlight regulations.

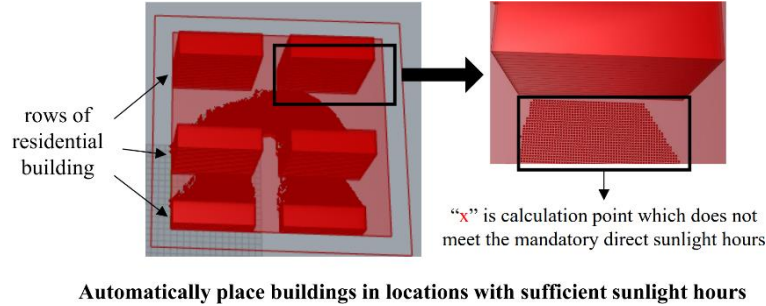

Fig. 18 Automatic residential neighborhood layout planning plug-in based on the MLP model

### 6.2. Discussion

The limitations of our approach are that it may only apply for assessment under the Chinese sunlight policy and in the conceptual design phase. The Chinese sunlight assessment calculates sunlight hours in a specified day with the worst sunlight condition in a year; we cannot take advantage of the feature that a point is only shaded once by a building for annual or quarterly sunlight simulation, which means that it is difficult to propose such a simple ANN architecture to predict the sites' heatmaps in these cases. Moreover, our model addresses cuboid-form buildings, which limits its application to the conceptual design phase. Buildings with complex forms may not satisfy translation or scaling equivalence.

In the future, we would like to further optimize the proposed method to make it applicable in general sunlight simulation cases. For example, we will attempt quarterly or annual sunlight hour prediction and more complex architectural forms. We may further adopt the idea of divide and conquer and/or try novel coordinate transformation to reach these goals. On the other hand, we will use deep-learning techniques in other building simulation fields. We may propose deep-learning-based proxy models for simulations of the daylight performance or energy consumption of buildings.

## Acknowledgments

This study was supported by the Research Project of Beijing Municipal Science & Technology Commission, Administrative Commission of Zhongguancun Science Park (20220468132) and the Research Development Project of the Ministry of Housing and Urban–Rural Development of the People's Republic of China (K20210032). We would also like to sincerely express our appreciation to Prof. Lin Borong of Tsinghua University, China, for providing the MOOSAS simulator.

## Appendix A. Translation of sunlight regulations in Chinese standards

### A.1. Standard for assessment parameters of sunlight on building [8]

**Regulation 2. 0. 3**: "reference day of sunlight assessment" is the specified day to measure and calculate "sunlight duration time" on buildings.

**Regulation 2. 0. 4**: "period of effective sunlight" is the time interval determined by the sun altitude and azimuth, intensity of solar radiation of "reference day of sunlight assessment", expressed in terms of apparent solar time.

**Regulation 2. 0. 5**: "reference position for sunlight assessment" is the calculation position on corresponding buildings and site for normalizing the calculation process of "sunlight duration time".

**Regulation 2. 0. 6**: "sunlight duration time" is the length of a continuous time interval or cumulative length of multiple time segments receiving sunlight at the "reference position for sunlight assessment" during the "period of effective sunlight".

**Regulation 2. 0. 7**: "standard of sunlight on buildings" means the minimum sunlight duration time of sunlight on buildings or site during the "period of effective sunlight" of "reference day of sunlight assessment", which is affected by the climate zone of the site, population of the city and the function of the buildings.

**Regulation 2. 0. 8**: "reference year of sunlight assessment" is the year to provide sun trajectory data for sunlight assessment.

**Regulation 5. 0. 1**: the setup of sunlight assessment should obey the following regulations: (1) the "reference year of sunlight assessment" should be 2001; (2) the scale of sampling grid should be reasonably determined according to the calculation method and the scope of calculation area, 0.3~0.6 meters for window, 0.6~1.0 meters for building, and 1.0~5.0 meters for site is appropriate; (3) it is inappropriate that time step more than 1 minute.

**Regulation 5. 0. 5**: Sunlight assessment should adopt apparent solar time, the time period can be counted cumulatively, and the minimum continuous segment that can be counted should not be less than 5 minutes.

**Explanation of Regulation 5. 0. 5**: Due to the increasing density of buildings in cities, a building may be shaded by multiple other buildings, its sunlight period is discontinuous, and a sunlight time segment may be a few minutes or less. Do not count segments less than 5 minutes due to the following reasons: (1) very short time segments have poor sunlight quality; (2) the bias between simulated and real "sunlight duration time" is usually 3~5 minutes.

### A.2. Standard for urban residential area planning and design [7]

**Regulation 4. 0. 9**: The spacing of residential buildings should follow the regulations in Table A.1 (major cold and winter solstice are specified date on the Chinese lunisolar calendar; they are January 20th and December 22nd in 2001, respectively; Shanghai belongs to the III climatic region [54]).

Table A.1 Standard of sunlight on residential buildings

| Climatic region for architecture | I, II, III, VII | | IV | | V, VI |
|---|---|---|---|---|---|
| Population of the city (thousand) | ≥ 500 | < 500 | ≥ 500 | < 500 | unlimited |
| Reference day of sunlight assessment | major cold | | | | winter solstice |
| Sunlight duration time (hour) | ≥ 2 | | ≥ 3 | | ≥ 1 |
| Period of effective sunlight (apparent solar time) | 8:00~16:00 | | | | 9:00~15:00 |
| Reference position for sunlight assessment | the plane of windowsill on the first floor | | | | |

\* the plane of windowsill on the first floor means the outer walls 0.9 meters above the ground of the first floor

## Appendix B. Validation of our SC-based sunlight simulator

We developed an SC-based sunlight simulator, which is used to train and validate our MLP-based model. The accuracy of our SC-based simulator is validated by Ladybug [17], Open3D [16], and Glodon Sunlight Analysis Software [13]. Our SC-based sunlight simulator is also faster than others because of the following two reasons: (1) our simulator is specified for cuboid-form buildings and gridded sampling points, and we conduct scenario-dependent code optimization; (2) our simulator adopts a scan line algorithm [27, 28], while others adopt a ray casting algorithm [31].

The results of the accuracy comparisons are listed in Table B.1. We first build a node graph with Ladybug in Grasshopper, as shown in Fig. B.1, and export its results for comparison; Fig. B.2 shows an example of the heatmap calculated by Ladybug and our SC-based simulator. We next compared our simulator with Open3D, which is an open sourced software. Glodon Sunlight Analysis Software was developed by another team of our company, and they finally provided us with an API for the comparison.

Table B. The results of accuracy comparison

| Compared software | MAE | Remark |
|---|---|---|
| Ladybug | 1.75 minutes | The sampling solar times are not exactly the same as ours |
| Open3D | 0.00 minutes | All inputs for simulation are exactly the same as ours |
| Glodon Sunlight Analysis Software | 1.72 minutes | The sampling points are not exactly the same as ours |

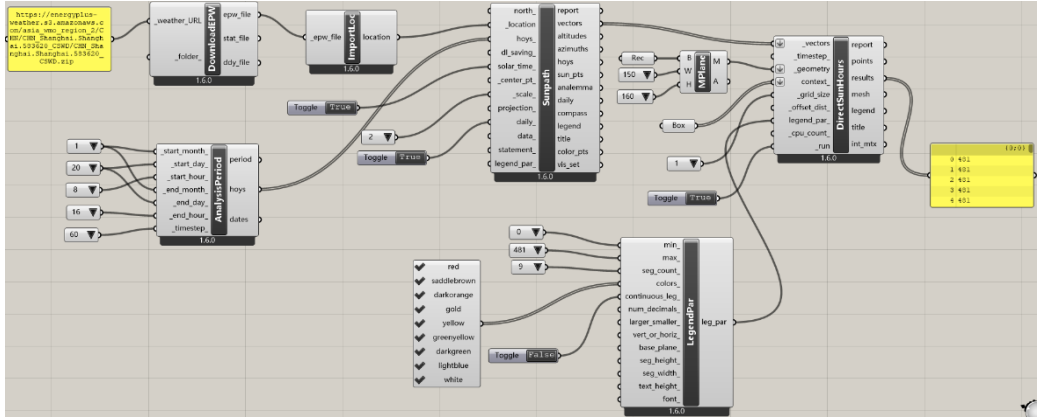

Fig. B.1 Node Graph for sunlight simulation in Ladybug/Grasshopper

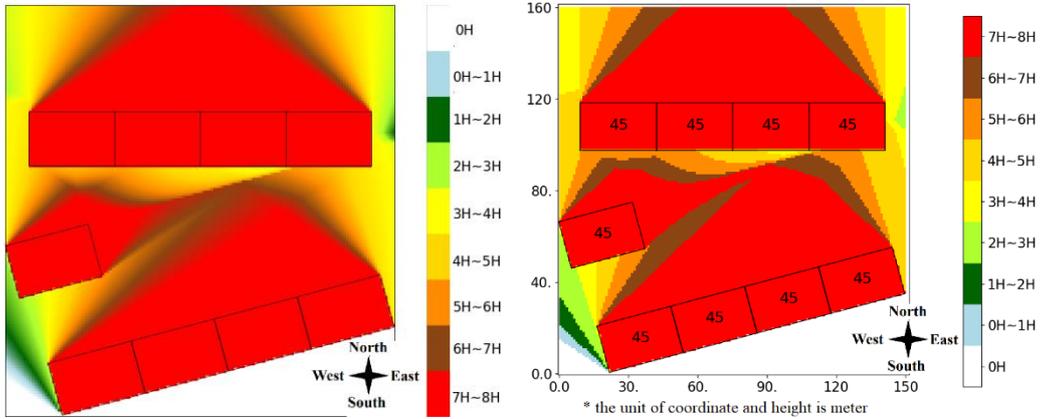

(a)                              (b)

* Ladybug's colormap is continuous, while ours is discrete, which makes the rendered images look slightly different

Fig. B.2 Total time spent shaded on the site calculated by (a) Ladybug and (b) our SC-based tool

We validate the calculation speed of our SC-based simulator by comparing it with others, as shown in Table B.2. The metric is computation time per time per sampling point. The result proves that our simulator is significantly faster than others. Moreover, the MLP-based prediction model is faster than our SC-based simulator, which shows its outstanding capabilities in terms of computation speed.

Table B.2 Comparison among our shadow analysis algorithms and others

| | Number of analysis times | Number of sampling points | Total computation time | Computation time per time per point |
|---|---|---|---|---|
| Ladybug | $4.81 \times 10^2$ | $4.8 \times 10^4$ | $2.06 \times 10^1$ s | $8.92 \times 10^{-7}$ s |
| Glodon sun analysis software | $4.8 \times 10^1$ | $1.12 \times 10^5$ | $2.61 \times 10^1$ s | $4.85 \times 10^{-6}$ s |
| Wang et al. [35] | $4.75 \times 10^3$ | $2.24 \times 10^3$ | $3.60 \times 10^4$ s | $3.38 \times 10^{-3}$ s |
| MOOSAS [55] | $1.2 \times 10^1$ | $2.76 \times 10^2$ | $1.55 \times 10^{-1}$ s | $4.68 \times 10^{-5}$ s |
| Open3D | $4.8 \times 10^2$ | $4.8 \times 10^4$ | $1.45 \times 10^0$ s | $6.29 \times 10^{-8}$ s |
| Optix | $4.8 \times 10^2$ | $4.8 \times 10^4$ | $9.31 \times 10^{-1}$ s | $4.04 \times 10^{-8}$ s |
| Ours SC-based tool | $4.8 \times 10^2$ | $4.8 \times 10^4$ | $3.08 \times 10^{-1}$ s | $1.34 \times 10^{-8}$ s |

## Appendix C. Proofs

### C.1. Proof: any point on the site is only shaded once by a cuboid-form building during the "period of effective sunlight"

It is difficult to prove this proposition analytically; thus, we indicate it with numerical experiments. The "period of effective sunlight" for Shanghai is 8:00~16:00 on January 20th, 2001 (AST) [7], and the size and orientation of buildings are defined in Table 4. We randomly generated 10000 scenarios and randomly sample points in the coarse grid (see Fig. 9). We conducted shading calculations for these points, and the delta time is 1 minute. Only 177 out of 19200000 points are shaded by a cuboid-form building more than once, i.e., the incidence is less than 0.001%.

### C.2. Proof: Correctness of the fast union method described in Table 3

Define that a point shaded by a building is an event, and the sunlight hours are the total time of at least one event happening, which can be calculated by Eq. (C. 1).

$$\text{time of at least one event happen} = \text{time when last event end} - \text{time when first event start} - \text{time of no event happen} \quad \text{(C. 1)}$$

Fig. C.1 shows how to calculate the total time of no event happening. We need to first sort all the start and end times. $ST_i$ and $ET_i$ represent the $i^{th}$ start and end event; although they are usually different events (the 2nd start event is #C, while the 2nd end event is #B in Fig. C.1), $ET_i$ is always later than $ST_i$, which can be proven by mathematical induction. If $ST_{i+1} > ET_i$, there is a time interval of no event happening, and its length is $ET_i - ST_{i+1}$ (the 3rd start event is later than the 2nd end event in Fig. C.1). If $ST_{i+1} \le ET_i$, all times between $ST_i$ and $ST_{i+1}$, at least one event happens (all times between $ST_1$ and $ET_2$, at least one event happens in Fig. C.1).

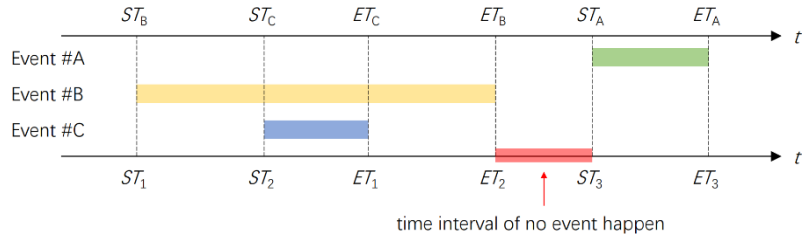

Fig. C.1 Proof of the fast union method

### C.3. Proof: Shadows of buildings are almost always in the valid sampling area under the setup in Section 4

We also randomly generated 10000 scenarios and found that 4525061 out of 4765438 shaded points are in the valid sampling area, i.e., the ratio is approximately 95%.

## Appendix D. Technical details

### D.1. Fast analytical method for shadow area calculation

Fig. D.1 illustrates a fast analytical method to calculate the shadow area of a cuboid-form building under a specified solar position. The altitude and azimuth of the sun are *alt* and *azi*, respectively (the positive azimuth direction is east). The height of the building is $H$, and its bottom section is ABCD, as shown in Fig. D.1 (a). We first translate ABCD $H \cdot \cot(alt) \cdot \cos(azi)$ westward and $H \cdot \cot(alt) \cdot \sin(azi)$ northward (sun is always in the northern sky during the calculation period required by the Chinese standard [7]), and A'B'C'D' is obtained as shown in Fig. D.1 (a). We then connect AA', BB', CC', and DD', and the shadow area is the purple shaded area in Fig. D.1 (b).

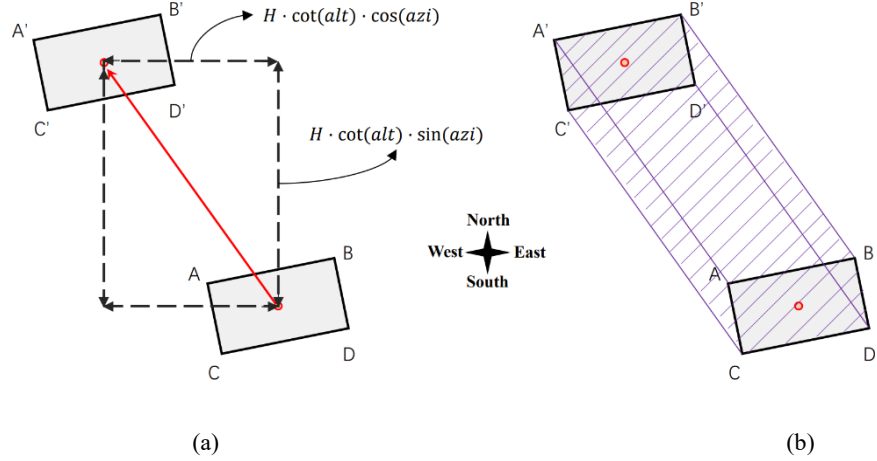

Fig. D.1 Fast analytical method to calculate shadow area: (a) building section translation; (b) shadow area

## D.2. Coordinate transformation before 3D point sunlight prediction

Assume that the length, width, height, and orientation of building **A** are $L$, $W$, $H$, and $\varphi$, respectively, and the latitude of the building is *lat*. To calculate its sunlight time interval on the point $(X, Y, 0)$, which means that the points are $X$ meters east and $Y$ meters north of the center of the building and are on the ground, we first need to calculate the relative length, width, and height based on Eqs. (4) to (7),

$$S = L + W + H \tag{D.1}$$

$$r_L = \frac{L}{S} \tag{D.2}$$

$$r_W = \frac{W}{S} \tag{D.3}$$

$$r_H = \frac{H}{S} \tag{D.4}$$

and the local Cartesian and polar coordinates of $(X, Y, 0)$ can be calculated according to the definitions (see Fig. 7).

$$r_x = \frac{X}{S} \tag{C.5}$$

$$r_y = \frac{Y}{S} + 0.2 \tag{C.6}$$

$$r_\theta = \arctan(\frac{r_y}{r_x}) \tag{C.7}$$

$$r_r = \sqrt{r_x^2 + r_y^2} \tag{C.8}$$

The sunlight time interval can be calculated by calling MLP($lat$, $r_L$, $r_W$, $r_H$, $\varphi$, $r_\theta$, $r_r$).

According to the translation equivalence discussed in Fig. 4 (a), the sunlight time interval on point

$(X, Y, Z)$, i.e., a point $Z$ meters above $(X, Y, 0)$, caused by building **A** equals the sunlight time interval on $(X, Y, 0)$ caused by building **B**, whose length, width, height, and orientation are $L$, $W$, $H - Z$, and $\varphi$, respectively. We can also calculate the relative length, width, and height of building **B** based on Eqs. (4) to (7),

$$S' = L + W + H - Z = \frac{1}{\eta}S \tag{D. 9}$$

$$r_L{}' = \frac{L}{S'} = \eta r_L \tag{D. 10}$$

$$r_W{}' = \frac{W}{S'} = \eta r_W \tag{D. 11}$$

$$r_H{}' = \frac{H - Z}{S'} = \eta(r_H - \frac{Z}{S}) \tag{D. 12}$$

where the coefficient $\eta$ is illustrated in Eq. (D. 13)

$$\eta = \frac{L + W + H}{L + W + H - Z} = 1 + \frac{Z}{S'} \tag{D. 13}$$

and the local Cartesian and polar coordinates can be revised accordingly,

$$r_x{}' = \frac{X}{S'} = \eta r_x \tag{D. 14}$$

$$r_y{}' = \frac{Y}{S'} + 0.2 = \eta r_y + 0.2(1 - \eta) \tag{D. 15}$$

$$r_\theta{}' = \arctan(\frac{r_y{}'}{r_x{}'}) = \arctan(\frac{r_y}{r_x} - \frac{0.2Z}{X}) \tag{D. 16}$$

$$r_r{}' = \sqrt{r_x'^2 + r_y'^2} \tag{D. 17}$$

The sunlight time interval can be calculated by calling MLP($lat$, $r`_L$, $r`_W$, $r`_H$, $\varphi$, $r`_\theta$, $r`_r$).

### D.3. Detailed objective function in the case of simulation-based optimization

According to Eq. (11), the objective score equals the total floor area score multiplied by the sunlight hour score, where the expression of the total floor area score is shown in Eq. (D. 18).

$$\text{total floor area score} = \frac{L \times W \times H}{42 \times 30 \times 60} \tag{D. 18}$$

where $L$, $W$, and $H$ are the length, width, and height of the placed building, respectively, and 42, 30, and 60 meters are the maximum values of $L$, $W$, and $H$, respectively. Note that the total floor area of the building is proportional to its volume if the floor height is a constant.

The sunlight hour score equals the product of two scores, as shown in Eq. (D. 19),

$$\text{direct sunlight hour score} = \min(1^{\text{st}} \text{ sunlight score}, 2) \times \min(2^{\text{nd}} \text{ sunlight score}, 2) \tag{D. 19}$$

where the $1^{\text{st}}$ and $2^{\text{nd}}$ sunlight scores correspond to the placed building and the three most northwest buildings in Fig. 15. We need to reduce the hours of the placed building shaded by the south buildings and the north buildings shaded by the placed one simultaneously. The sunlight scores are defined in Eq. (D. 20),

$$1^{\text{st}}/2^{\text{nd}} \text{ sunlight score} = \begin{cases} 0, \text{direct-sunligt-hours of any check point less than 1.75 hours} \\ \dfrac{\displaystyle\sum_{\text{check points}} \dfrac{\text{direct-sunligt-hours} - 2 \text{ hours}}{0.25 \text{ hours}}}{\text{number of check points}}, \text{ elsewise} \end{cases} \tag{D. 20}$$

where the check points are the sampling points outside the buildings and closest to their south facades.